\documentclass[11pt]{article}

\usepackage{acl}
\usepackage{times}
\usepackage{latexsym}
\usepackage[T1]{fontenc}
\usepackage[utf8]{inputenc}
\usepackage{microtype}
\usepackage{inconsolata}
\usepackage{graphicx}
\usepackage{amsmath}
\usepackage{amssymb}
\usepackage{tabularx}
\usepackage{listings}
\usepackage{xcolor}
\usepackage{algorithm}
\usepackage{algorithmic}

\definecolor{darkgreen}{rgb}{0.0, 0.5, 0.0}
\definecolor{codegreen}{rgb}{0,0.6,0}
\definecolor{codegray}{rgb}{0.5,0.5,0.5}
\definecolor{codepurple}{rgb}{0.58,0,0.82}
\definecolor{backcolour}{rgb}{0.98,0.98,0.98}

\usepackage{booktabs}
\definecolor{codegreen}{rgb}{0,0.6,0}
\definecolor{codegray}{rgb}{0.5,0.5,0.5}
\definecolor{codepurple}{rgb}{0.58,0,0.82}
\definecolor{backcolour}{rgb}{0.98,0.98,0.98}

\lstdefinestyle{pythonstyle}{
    backgroundcolor=\color{backcolour},
    commentstyle=\color{codegreen},
    keywordstyle=\color{magenta},
    numberstyle=\tiny\color{codegray},
    stringstyle=\color{codepurple},
    basicstyle=\footnotesize\ttfamily,
    breakatwhitespace=false,
    breaklines=true,
    captionpos=b,
    keepspaces=true,
    numbers=left,
    numbersep=6pt,
    showstringspaces=false,
    tabsize=2,
    language=Python,
}

\title{ToolBrain: A Flexible Reinforcement Learning Framework for \\ Agentic Tools}

\author{
    \mdseries 
    Quy Minh Le\textsuperscript{1}\thanks{Equal contribution.} \quad
    Minh Sao Khue Luu\textsuperscript{1}\footnotemark[1] \quad
    Khanh-Tung Tran\textsuperscript{2}\footnotemark[1] \quad
    Duc-Hai Nguyen\textsuperscript{2} \\
    Hoang-Quoc-Viet Pham\textsuperscript{1} \quad
    Quan Le\textsuperscript{3} \quad
    Hoang Thanh Lam\textsuperscript{4}\thanks{\raggedright Corresponding authors: \texttt{t.l.hoang@ie.ibm.com}, \texttt{hn@cs.ucc.ie}.} \quad
    Hoang D. Nguyen\textsuperscript{2}\footnotemark[2]
    \\[1.0ex]
    \textsuperscript{1}ToolBrain Research, Ireland \\
    \textsuperscript{2}University College Cork, Ireland \\
    \textsuperscript{3}CeADAR University College Dublin, Ireland \\
    \textsuperscript{4}IBM Research Lab, Dublin, Ireland
}

\begin{document}
\maketitle
\begin{abstract}
Effective tool use is essential for agentic AI, yet training agents to utilize tools remains challenging due to manually designed rewards, limited training data, and poor multi-tool selection, resulting in slow adaptation, wasted computational resources, and suboptimal performance. We introduce \emph{ToolBrain}, a lightweight and user-friendly framework for training tool use in agentic models with flexible reinforcement learning, thereby easing the barriers for researchers and practitioners to adapt LLM-based agents to specific domains. It supports a wide range of training strategies, including reinforcement learning algorithms such as GRPO and DPO, as well as supervised learning. ToolBrain enables custom reward callables directly on an agent’s execution traces or simply utilizes an automated LLM-as-a-judge system for reward generation. It is packed with useful capabilities, including knowledge distillation from large to small models, automatic task generation from tool descriptions, seamless tool retrieval, efficient fine-tuning pipelines with QLoRA through Unsloth, and quantized inference via bitsandbytes. We demonstrate ToolBrain through an Email Search Agent case study, showing measurable improvements in tool-use skills under a realistic workflow, while keeping the codebase simple and extensible. Our framework is publicly available at https://toolbrain.org/
\end{abstract}

\section{Introduction}

LLM-based agents have become a viable technological wave, capable of executing complex tasks, ranging from planning, code generation, interaction with APIs, to scientific discovery, through the use of tools \cite{shinn2023reflexion, yao2022react, schick2023toolformer, langgraph2025}.

However, many agentic systems rely on supervised fine-tuning or prompt engineering to adapt behavior \cite{openai2023gpt4, wei2022chain}, preventing continuous improvements through experience in perplexing environments. Reinforcement Learning (RL) enables adaptive policies based on system traces, enhancing LLM capabilities for preference optimization \cite{ouyang2022training, rafailov_direct_2024} and reasoning tasks \cite{nakano2021webgpt}, but its integration into agentic tool-use workflows remains underdeveloped, especially for complex and self-evolving tools.

Several key challenges hinder the broader adoption of RL in agent and tool development. First, existing frameworks such as ART \cite{hilton2025art} and Agent Lightning \cite{luo2025agentlightningtrainai} lack a lightweight and user-friendly interface that allows users to define and apply RL reward signals directly on an agent execution trace. Second, although tool calling with large language models is highly effective, these models remain computationally expensive, while smaller models perform substantially worse. As a result, knowledge distillation from large models becomes critical for industrial deployment and cost efficiency. Third, the tool ecosystem is often extremely large, making it inefficient to learn effective behavior in the presence of many irrelevant tools. Finally, collecting high-quality training data is typically very costly. Addressing the challenge of teaching models to use tools effectively, therefore, requires treating all of these issues in a unified manner.

ToolBrain is designed to address these pain points by providing a simple API that connects (i) a tool-using agent, (ii) a flexible reward system (\S\ref{sec:feature-rewards}), and (iii) RL training algorithms such as GRPO and DPO (\S\ref{sec:feature-algorithms}). The framework is further extended with a suite of powerful features, including intelligent tool retrieval (\S\ref{sec:feature-retrieval}), zero-shot task generation (\S\ref{sec:feature-zero-learn}), knowledge distillation (\S\ref{sec:feature-distillation}), and a highly efficient training backend (\S\ref{sec:feature-efficiency}).

We demonstrate ToolBrain's effectiveness and flexibility through a comprehensive evaluation across multiple tasks. Our main experiment focuses on a complex, multi-step information retrieval task (Email Search), where we show significant performance improvements on both 3B and 7B parameter models. To showcase the framework's rapid adaptation capabilities, we conduct two supplementary experiments on quantitative reasoning (Finance) and real-world grounding (API) tasks, demonstrating how Knowledge Distillation can quickly uplift the performance of a compact 0.5B model.

The remainder of this paper is structured as follows. We first situate our work within the existing landscape of agent frameworks (\S\ref{sec:related_work}). We then describe the technical details of ToolBrain, presenting our core Coach-Athlete architectural paradigm and its key features (\S\ref{sec:system}). Subsequently, we provide a comprehensive empirical evaluation to validate our framework's performance on the aforementioned tasks (\S\ref{sec:experiments}). Finally, we conclude with a summary of our contributions (\S\ref{sec:conclusion}).

\section{Related Work}
\label{sec:related_work}

\begin{table*}[t]
  \centering
  \renewcommand{\arraystretch}{1.25}
  \small
  \begin{tabularx}{\textwidth}{
    >{\raggedright\arraybackslash}p{2.6cm}
    >{\raggedright\arraybackslash}X
    >{\raggedright\arraybackslash}X
    >{\raggedright\arraybackslash}X
    >{\raggedright\arraybackslash}X
  }
    \hline
    \textbf{Aspect} &
    \textbf{ToolBrain} &
    \textbf{LangChain / LangGraph} &
    \textbf{ART} &
    \textbf{Agent Lightning} \\
    \hline

    Training Approach &
    \textcolor{darkgreen}{\checkmark}\ \textbf{Native RL (GRPO, DPO)} with iterative fine-tuning. &
    Supervised learning and prompt chaining. &
    GRPO-based RL with RULER evaluator. &
    Hierarchical RL with credit assignment. \\

    Reward System &
    \textcolor{darkgreen}{\checkmark}\ \textbf{Hybrid:} Python callable plus ranking-based LLM. &
    Manual heuristic scoring. &
    RULER: automated LLM-as-judge with relative scoring. &
    Credit-aware reward assignment. \\

    Tool Management &
    \textcolor{darkgreen}{\checkmark}\ \textbf{Integrated Tool Retriever} automatically selects relevant tools. &
    Manual tool definition and passing. &
    Manual tool definition and passing. &
    Manual tool definition and passing. \\

    Advanced Strategies &
    \textcolor{darkgreen}{\checkmark}\ Supports \textbf{knowledge distillation} and \textbf{Zero-Learn} task generation. &
    --- &
    --- &
    --- \\

    Efficiency \& Usability &
    \textcolor{darkgreen}{\checkmark}\ \textbf{Simple \texttt{Brain} API}; integrated Unsloth/QLoRA optimizations. &
    Code-centric; complex context management. &
    Minimal code changes; requires separate server setup. &
    Requires MDP design; steep RL expertise. \\
    \hline
  \end{tabularx}
  \caption{Comparison of ToolBrain with other agent training frameworks.}
  \label{tab:related_work}
\end{table*}

ToolBrain builds upon a rich landscape of agent and RL frameworks.
Although numerous systems facilitate agent development, they often
present trade-offs in usability, flexibility, and the steep learning curve associated with reinforcement learning. To position our contributions, we compare ToolBrain with three representative approaches in Table~\ref{tab:related_work}: \textit{LangChain/LangGraph}\cite{langgraph2025}, representing popular code-centric systems; \textit{ART}\cite{hilton2025art}, a contemporary RL-focused framework; and \textit{Agent Lightning}\cite{luo2025agentlightningtrainai}, which focuses on hierarchical RL. ToolBrain aims to complement these systems by offering a lightweight, RL-centric interface that exposes reward signals at the level of agent traces and integrates practical features such as distillation and tool retrieval.

\section{System Description}
\label{sec:system}

\subsection{Architecture Overview}

\begin{figure}[t]
    \centering
    \includegraphics[width=\columnwidth]{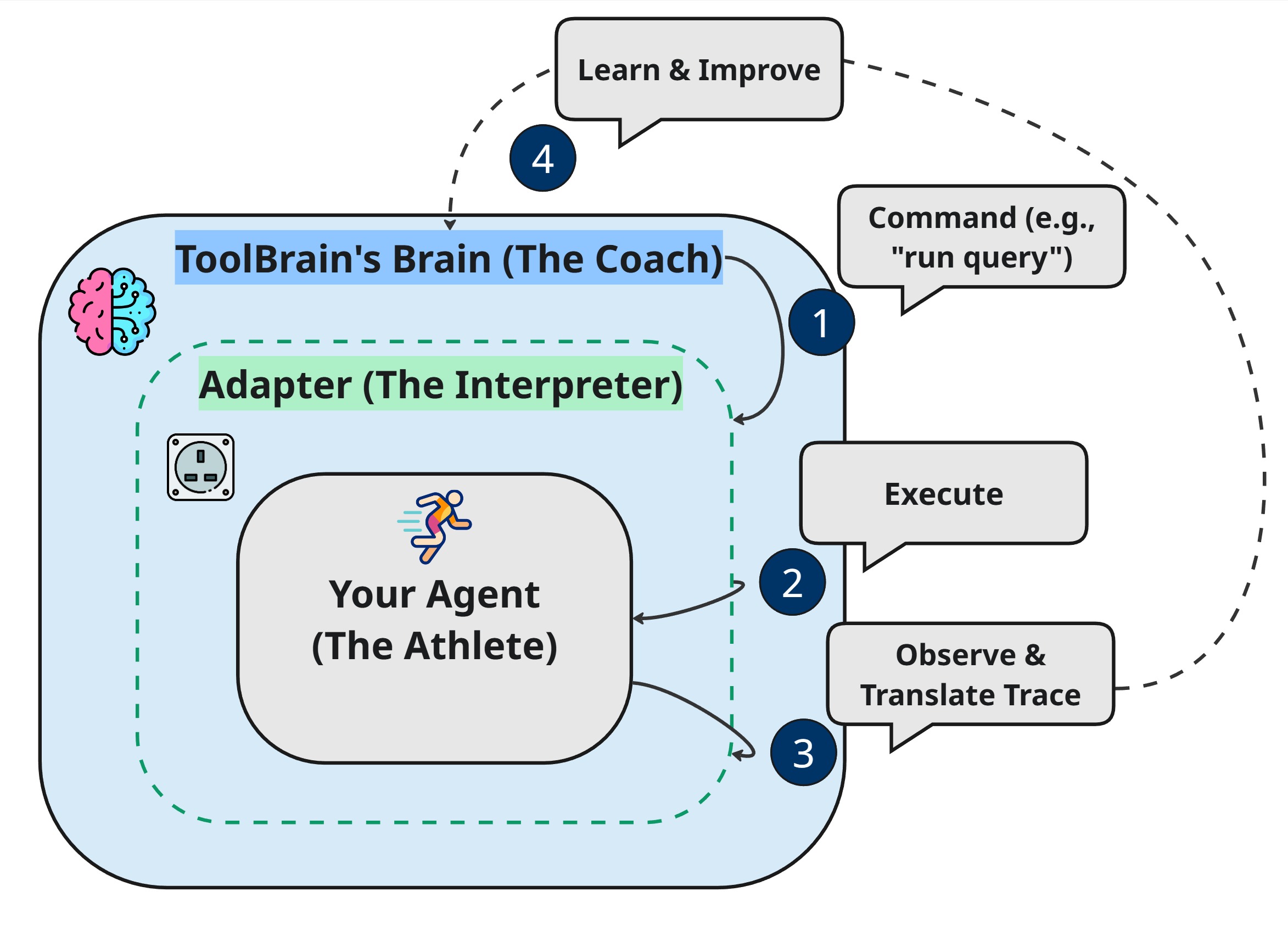}
    \caption{The Coach-Athlete-Interpreter paradigm in ToolBrain. The Brain orchestrates the process, the user-provided Agent executes the task, and the Adapter acts as a standardized communication layer, translating interactions into a high-fidelity trace for learning.}
    \label{fig:paradigm}
\end{figure}

ToolBrain’s design is structured around the \textbf{Coach–Athlete paradigm}, a conceptual abstraction that cleanly separates \emph{training orchestration} from \emph{task execution responsibilities}. As illustrated in Figure~\ref{fig:paradigm}, the paradigm consists of three main components: a high-level \textbf{Brain} (the Coach) that manages the training loop; an \textbf{Agent} (the Athlete) responsible for executing tasks using external tools; and a lightweight internal \textbf{Adapter} (the Interpreter) that provides a standardized communication interface.

The training workflow progresses through a complete, well-defined cycle. First, the Brain issues a command to the Agent. The Agent executes the task, and its actions are monitored by the Adapter, which translates them into a standardized, high-fidelity \texttt{Execution Trace}. This trace is the foundation of the learning loop: the Brain uses it to compute rewards and determine the optimal policy updates. Crucially, the Brain then applies these updates directly to the parameters of the Agent's underlying model, thereby improving its capabilities for future tasks and completing the cycle.

Although this paradigm shares conceptual similarities with actor-learner architectures such as IMPALA \citep{impala2018}, it establishes a different boundary tailored for iterative agent development. The linchpin of this architecture is the internal Adapter, which implements the classic Adapter design pattern \citep{gamma1995design}. Its primary responsibility is to act as a structural bridge, converting the proprietary, framework-specific memory of a user-provided agent into ToolBrain's standardized \texttt{Execution Trace} format.

This decoupling is critical: it allows the Brain (the client) to interact with any Agent (the adaptee) through a single, consistent interface, regardless of the agent's internal implementation (e.g., \texttt{smolagents}, \texttt{langchain}). By handling this translation automatically, the Adapter makes the framework agnostic to the user's choice of agent implementation. This ensures that the trace data fed to downstream RL algorithms is uniform and high-fidelity, regardless of the underlying agent's architecture. Comprehensive schema definitions for the \texttt{Execution Trace} are provided in Appendix~\ref{sec:impl-details}.

\begin{figure*}[t]
    \centering
    \includegraphics[width=0.95\textwidth]{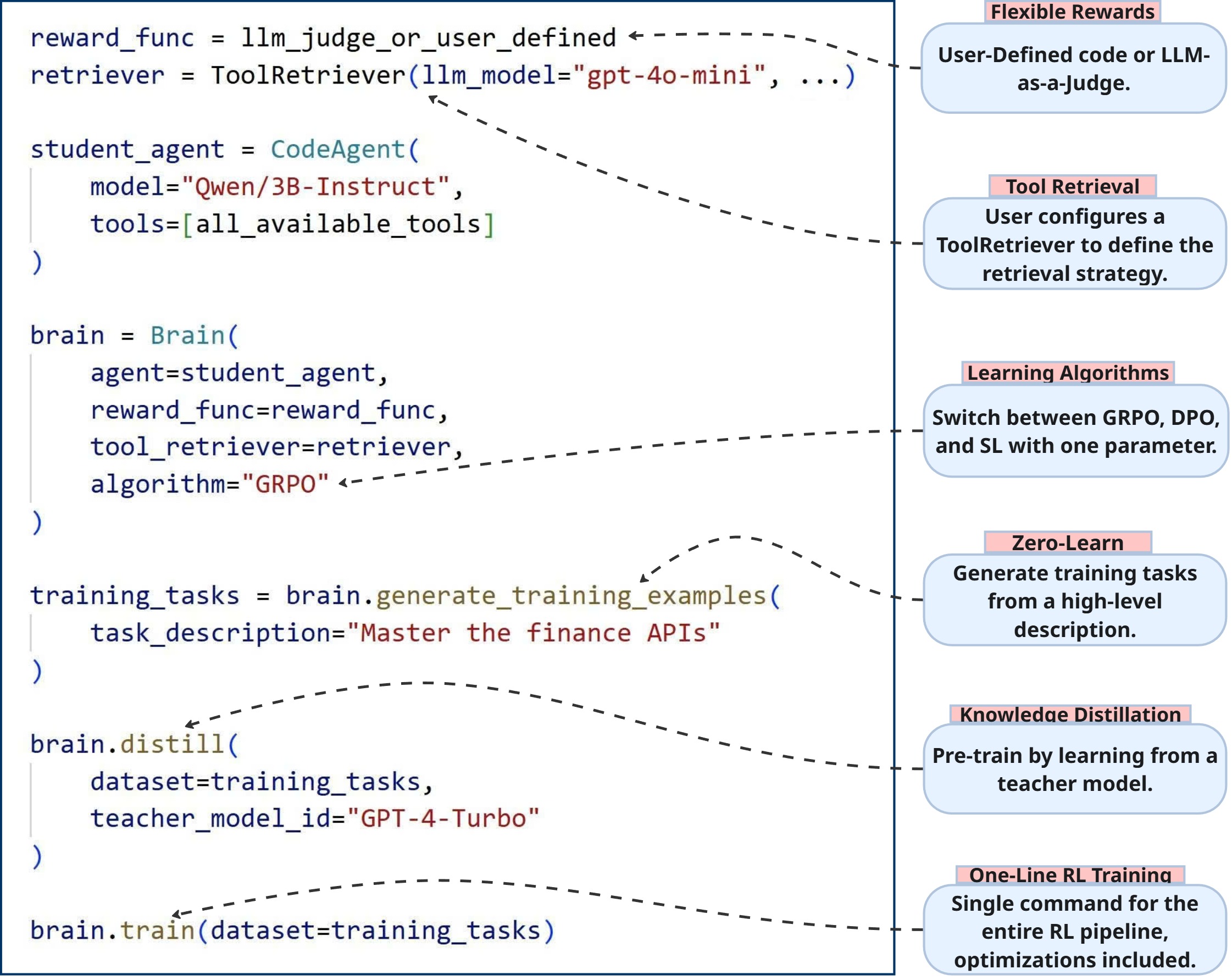}
    \caption{The ToolBrain API workflow, demonstrating how a user composes a training setup and executes the entire RL pipeline with a single command. It highlights key features such as the configurable tool retrieval, flexible rewards, one-parameter algorithm selection, and advanced strategies like knowledge distillation.}
    \label{fig:code_diagram}
\end{figure*}

Figure~\ref{fig:code_diagram} illustrates the streamlined API workflow of ToolBrain, which encapsulates the entire training process. The design emphasizes modularity and user control. The user first configures the core components: a tool-using \texttt{agent}, a \texttt{reward\_func}, and optionally a custom \texttt{ToolRetriever} to define the tool selection strategy. These components are then passed to the central Brain orchestrator, which manages the entire pipeline~---~from data generation and knowledge distillation to the final one-line \texttt{brain.train()} command.

\subsection{Key Features and Innovations}
\label{sec:key-features}

ToolBrain introduces several features intended to make RL for tool-using agents more accessible in practice.

\subsubsection{Flexible Reward Interface}
\label{sec:feature-rewards}
A central goal of ToolBrain is to simplify reward design, which remains a significant challenge in agent training. To this end, our framework provides a flexible, hybrid reward system that supports two complementary approaches. The first is user-defined heuristic rewards, where any Python callable can operate directly on an agent's full execution trace. This allows for precise, objective, and domain-specific feedback based on clear success criteria.

For more nuanced tasks without clear ground-truth, ToolBrain integrates an optional LLM-as-a-Judge mechanism. This aligns with the growing body of work on using powerful LLMs for evaluation, a practice increasingly validated in recent research \citep{zheng2023judging, liu2023geval, gu2025surveyllmasajudge}. Instead of asking for an unreliable absolute score, ToolBrain's judge ranks a group of traces for the same query from best to worst. This relative feedback is then converted into normalized scalar rewards, a robust approach inspired by the principles of learning from human preferences \citep{ouyang2022training}.

Internally, a small wrapper harmonizes both heuristic and judge-based functions into a unified interface, abstracting away the complexity from the training loop. Implementation details and example reward functions are provided in Appendix~\ref{sec:reward-appendix}.

\subsubsection{Learning Algorithms: GRPO and DPO}
\label{sec:feature-algorithms}
ToolBrain supports two state-of-the-art algorithms for agent alignment. For scenarios with explicit scalar rewards, it implements Group Relative Policy Optimization (GRPO) \citep{shao_deepseekmath_2024}. GRPO is a modern variant of the widely-used Proximal Policy Optimization (PPO) family of algorithms \citep{schulman2017proximal}, which optimizes a policy by generating groups of responses and normalizing rewards to obtain stable relative advantages.

For preference-based learning, ToolBrain implements Direct Preference Optimization (DPO) \citep{rafailov_direct_2024}. DPO offers a more direct and often more stable approach to the problem of learning from human feedback \citep{christiano2017deep}, as it learns directly from chosen vs.\ rejected pairs without needing to train a separate reward model. We provide full pseudocode for both methods in Appendix~\ref{sec:rl-algorithms}.

\subsubsection{Intelligent Tool Retrieval}
\label{sec:feature-retrieval}
As the number of available tools grows, an agent's ability to select the most relevant ones for a given task becomes critical. Providing an LLM with a vast and mostly irrelevant tool library increases context length, computational cost, and the likelihood of hallucinated or incorrect tool calls. This challenge is a central theme in recent tool-use research \citep{schick2023toolformer, qin2023toolllm}.

To address this, ToolBrain implements an Intelligent Tool Retrieval mechanism. Our approach externalizes the selection logic from the agent itself, a design inspired by recent systems that employ a dedicated LLM as a tool filter or retriever \citep{qin2023toolllm, Huang2025Biomni}. Instead of a simple boolean flag, ToolBrain uses a dependency injection pattern: the user instantiates and configures a separate \texttt{ToolRetriever} object, which leverages a powerful LLM (e.g., GPT-4) to act as a tool selection module. This retriever is then passed to the Brain during initialization.

Before each task, the Brain uses this retriever to dynamically select a small, relevant subset of tools from the agent's full library. This design provides significant flexibility, reduces context length, and improves agent accuracy by focusing its attention on the most pertinent tools for the task at hand. An example is provided in Appendix~\ref{sec:tools-efficiency}.

\subsubsection{Zero-Learn Task Generation}
\label{sec:feature-zero-learn}
A primary obstacle in fine-tuning agents is the scarcity of high-quality, domain-specific training data. To address this, ToolBrain incorporates a Zero-Learn task generation pipeline, enabling it to bootstrap its own training data. This approach is inspired by a growing body of work demonstrating that large language models can effectively self-improve by learning from their own generated outputs \citep{wang2023self, huang2022large}.

Given a set of tool definitions and an optional high-level task description, ToolBrain's \texttt{generate\_training\_examples} method prompts its own underlying LLM to synthesize a multiple set of tool-using queries. This self-supervised methodology aligns with the principles of models that teach themselves to use tools \citep{schick2023toolformer}. To further enhance data quality, the framework includes a self-ranking mechanism, allowing the model to evaluate its own generations and prioritize tasks that are concrete and well-aligned with the provided tools. A detailed breakdown of the generation process and query categories is provided in Appendix~\ref{sec:zero-learn}.

\subsubsection{Knowledge Distillation for Policy Initialization}
\label{sec:feature-distillation}
Training smaller, more efficient models to perform complex tasks is a primary goal in applied NLP. To facilitate this, ToolBrain implements a knowledge distillation pipeline, a powerful technique for transferring capabilities from a large \textit{teacher} model to a smaller \textit{student model} \citep{hinton2015distilling}. This approach has proven highly effective for creating compact and fast language models, such as DistilBERT, without a significant loss in performance \citep{sanh2020distilbert}.

In ToolBrain's workflow, a large teacher model (e.g., 7B parameters) first generates high-quality execution traces for a given set of tasks. These expert demonstrations are then filtered for correctness and used as a dataset to train a smaller student model (e.g., 0.5B parameters) via a standard supervised learning objective. This process provides the student with a well-initialized policy, yielding meaningful action distributions and stable behavior prior to subsequent reinforcement learning fine-tuning. As demonstrated in our supplementary experiments (Section~\ref{sec:experiments}), this technique significantly uplifts the performance of small models on specialized tasks. The full algorithm and code examples are provided in Appendix~\ref{sec:distillation}.

\subsubsection{Efficient Training Optimization}
\label{sec:feature-efficiency}
Making RL fine-tuning practical on consumer hardware is a core design principle of ToolBrain. To achieve this, the framework integrates a suite of state-of-the-art optimization techniques, abstracting them behind simple, high-level parameters.

A key strategy is the support for Low-Rank Adaptation (LoRA) \citep{hu2021lora}, which dramatically reduces the number of trainable parameters by freezing the base model and only training small, injectable matrices. Building upon this, ToolBrain natively supports QLoRA \citep{dettmers2023qlora}, an even more memory-efficient approach that uses 4-bit quantization for the frozen base model, making it feasible to fine-tune large models on a single GPU.

Furthermore, the framework supports standard mixed-precision training (fp16) to accelerate computation \citep{micikevicius2018mixed}, and seamlessly integrates with accelerated backends such as Unsloth \citep{unsloth} to improve training efficiency. By combining these powerful optimizations, ToolBrain significantly lowers the barrier to entry for training capable, tool-using agents. Implementation examples are provided in Appendix~\ref{sec:training-opts-appendix}.

\section{Experiments and Results}
\label{sec:experiments}

To demonstrate the effectiveness and flexibility of ToolBrain, we conduct a comprehensive evaluation centered on three distinct tasks. Our main experiment provides a deep dive into a complex, multi-step reasoning task, showcasing the framework's core training capabilities. This is followed by two additional experiments that assess the framework’s versatility and efficiency in adapting agents to specialized domains.

\subsection{Main Experiment: Email Search Agent}

\subsubsection{Task}
This experiment centers on a challenging information retrieval and multi-step reasoning task, inspired by the ART$\cdot$E project \citep{arte_blog_post}. The agent's objective is to answer natural language questions by navigating a large-scale email database. We use the well-established Enron Email Corpus \citep{klimt2004enron}, a dataset containing approximately 0.5 million real-world emails from senior management at Enron. For training and evaluation, we utilize a publicly available set of question-answer pairs derived from this corpus \citep{corbt_enron_questions_2025}. This setup requires the agent to perform a sophisticated tool-using workflow,
including searching, reading, and synthesizing information to derive the correct answer.

\subsubsection{Evaluation and Results}
We trained two sizes of the Qwen2.5 model (3B and 7B parameters) for 60 steps using the GRPO algorithm \cite{shao_deepseekmath_2024} with an LLM-as-a-Judge \cite{zheng2023judging, ouyang2022training}. To provide a comprehensive view of performance, we tracked multiple metrics: task success rate, hallucination rate (the percentage of incorrect answers among all attempted answers), and the average number of turns per query.

\begin{figure}[!ht]
    \centering
    \includegraphics[width=1.0\columnwidth]{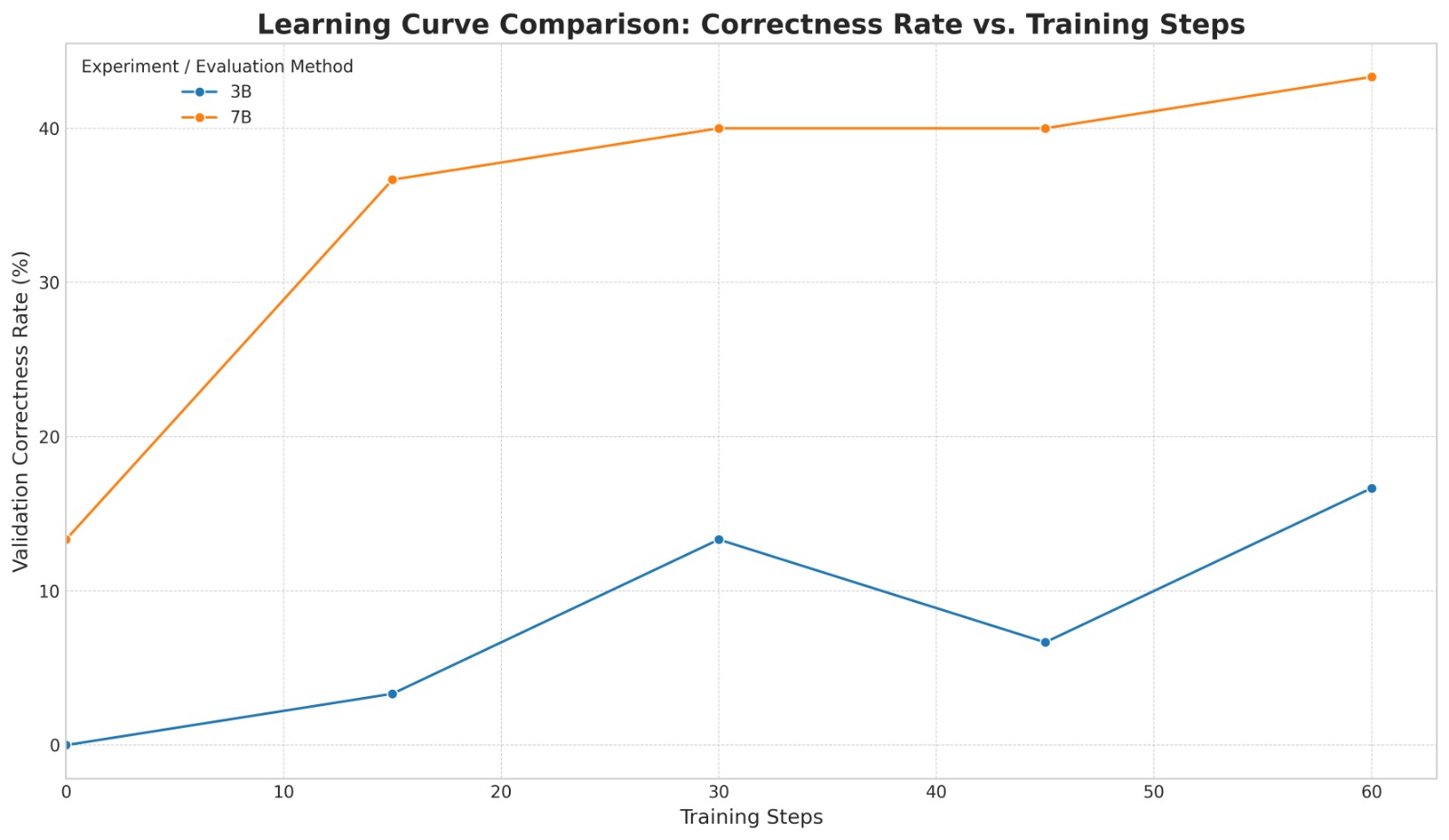}
    \caption{Learning curve showing the \textit{Correctness Rate} from a representative training run on the email search task.}
    \label{fig:learning_curve}
\end{figure}

The results summarized in Table~\ref{tab:email_results_summary} indicate a substantial improvement in the agent’s overall performance. The learning dynamics are further illustrated in Figure~\ref{fig:learning_curve}. The untrained 3B model fails to solve the task, whereas the 7B model demonstrates a small degree of zero-shot competence. After training, both models achieve notable gains in task success. Importantly, training enhances both accuracy and reliability, reducing the hallucination rate of the 7B model from 60.0\% to 35.0\%. Moreover, the 7B model shows improved efficiency, completing the task in fewer turns, while the 3B model typically engages in more turns to reach a solution. These contrasting behaviors appear to reflect the effects of scale on the two models. Finally, the continued upward trajectory of correctness for both models at the 60-step mark suggests that further training could yield additional improvements.

\begin{table*}[!ht]
\centering
\caption{Comprehensive Evaluation of the Email Search Agent (from a single representative run). The table compares key performance metrics at the start (Step 0) and end (Step 60) of training. Arrows (↓) indicate that lower values are better.}
\label{tab:email_results_summary}

\resizebox{\textwidth}{!}{
\begin{tabular}{lcccccc}
\toprule
& \multicolumn{3}{c}{\textbf{Before Training (Step 0)}} & \multicolumn{3}{c}{\textbf{After Training (Step 60)}} \\
\cmidrule(lr){2-4} \cmidrule(lr){5-7}
\textbf{Model} & \textbf{Success Rate (\%)} & \textbf{Hallucination Rate (\%)} ↓ & \textbf{Avg. Turns} ↓ & \textbf{Success Rate (\%)} & \textbf{Hallucination Rate (\%)} ↓ & \textbf{Avg. Turns} ↓ \\
\midrule
Qwen2.5-3B  & 0.0  & 100.0 & 4.63 & \textbf{16.7} & 66.7 & 5.57 \\
Qwen2.5-7B  & 13.3 & 60.0  & 7.03 & \textbf{43.3} & 35.0 & 4.77 \\
\bottomrule
\end{tabular}
}
\end{table*}

\subsection{Supplementary Experiments: Demonstrating Flexibility and Efficiency}

\subsubsection{Methodology and Tasks}
To showcase the framework's versatility, we evaluated a 0.5B parameter agent on two supplementary tasks, each representing a different challenge for language agents. We used a simple \texttt{Before vs. After} methodology to measure the impact of our training process. The tasks are as follows:
\begin{itemize}
    \item \textbf{Finance Agent (Quantitative Reasoning):} The agent must correctly map natural language questions to structured financial calculation tools, a task that requires high precision and strict adherence to tool specifications.
    \item \textbf{API Agent (Real-World Grounding):} The agent must call an external weather API to answer real-time queries, demonstrating its ability to connect to and utilize live, external data sources.
\end{itemize}

\subsubsection{Dataset and Training}
To prepare for training and evaluation, we used ToolBrain's built-in \texttt{generate\_training\_examples} feature to automatically synthesize datasets. For each of the two additional tasks, we generated a training set of 40 queries and a held-out test set of 10 queries by providing a task-specific description. The agents were trained briefly and efficiently using knowledge distillation from a 7B teacher model. This approach demonstrates the Zero-Learn capability of our framework while ensuring a consistent and reproducible methodology. Details on the data-generation API and the distillation pipeline are provided in Appendix~\ref{sec:zero-learn} and Appendix~\ref{sec:distillation}, respectively.

\subsubsection{Results and Analysis}
Evaluation was conducted on a held-out test set of 10 unseen queries for each agent. As shown in Table~\ref{tab:supplementary_results}, the untrained baseline models exhibited limited capability. After the distillation phase, the agents demonstrated clear improvements in calling the correct tool. The Finance Agent improved from 20\% to 40\%, while the API Agent improved from 30\% to 60\%. Although the results are not perfect, they indicate that ToolBrain~---~and especially its distillation component~---~offers an efficient mechanism for rapidly adapting small language models to specialized domains.

\begin{table}[!ht]
\centering
\caption{Demonstrating Framework Flexibility on Secondary Tasks. Success Rate (\%) is measured by the agent's ability to call the correct tool on a held-out test set of 10 queries.}
\label{tab:supplementary_results}
\begin{tabularx}{\columnwidth}{>{\raggedright\arraybackslash}X
                                    >{\centering\arraybackslash}p{2cm}
                                    >{\centering\arraybackslash}p{2cm}}
\toprule
\textbf{Case Study} & \textbf{Untrained} & \textbf{Trained (w/ Distill)} \\
\midrule
Finance (Quant. Reasoning)     & 20.0\% & 40.0\% \\
API (Real-World Grounding)     & 30.0\% & 60.0\% \\
\bottomrule
\end{tabularx}
\end{table}

\section{Conclusion}
\label{sec:conclusion}

The effective use of tools is fundamental to advancing the capabilities of agentic AI systems. However, developing robust tool-augmented agents presents several challenges, including the inherent complexity of reinforcement learning frameworks, the difficulties associated with designing appropriate reward functions, and the substantial computational cost required for training. In this work, we introduce ToolBrain, a framework that aims to bridge the gap between agent design and iterative, experience-driven improvement through reinforcement learning.

We introduce the \textit{Coach–Athlete paradigm} as a central architectural principle, offering a streamlined, high-level API that abstracts away underlying implementation complexity. We show how ToolBrain’s flexible hybrid reward system enables users to provide effective feedback through both user-defined code and a ranking-based LLM-as-a-Judge mechanism. In addition, we describe a set of advanced capabilities~---~such as intelligent tool retrieval, knowledge distillation, and zero-learn task generation~---~that operate alongside modern training optimizations, including Unsloth and QLoRA, to make agent training more practical and accessible.

Through our central case study of training an Email Search Agent, we provided both quantitative and qualitative evidence of our framework's efficacy. The experimental results show that agents trained with ToolBrain demonstrate significant and consistent performance improvements over their initial baselines. The learning curves validate our training pipeline's effectiveness, and the final agent's ability to handle complex, multi-step workflows showcases the sophisticated skills acquired.

Although ToolBrain provides a flexible foundation, several promising avenues remain for future work. These include expanding the tool retrieval mechanism to handle even more complex, dynamic tool libraries, extending the Coach-Athlete paradigm to multi-agent scenarios, and exploring online RL algorithms for continuous, real-time agent adaptation.

Ultimately, ToolBrain offers a practical and effective basis for the broader community. By reducing the barrier to agent-centric RL, we aim to enable more developers and researchers to design, refine, and deploy the next generation of capable, reliable, and domain-adapted autonomous systems.

\section*{Acknowledgements}
This publication has emanated from research supported in part by grants from Research Ireland under Grant [12-RC-2289-P2] and [18/CRT/6223] which is co-funded under the European Regional Development Fund. This work is also supported by ISY Labs. For the purpose of Open Access, the author has applied a CC BY public copyright licence to any Author Accepted Manuscript version arising from this submission.

\newpage
\bibliography{custom}

@article{ouyang2022training,
  title={Training language models to follow instructions with human feedback},
  author={Ouyang, Long and Wu, Jeff and Jiang, Xu and Almeida, Diogo and Wainwright, Carson and Mishkin, Pamela and Zhang, Chong and Agarwal, Sandhini and Slama, Katarina and Ray, Alex and others},
  journal={arXiv preprint arXiv:2203.02155},
  year={2022}
}

@article{nakano2021webgpt,
  title={WebGPT: Browser-assisted question-answering with human feedback},
  author={Nakano, Reiichiro and Hilton, Jacob and Balaji, Oleg and Chowdhery, Aakanksha and Hashme, Christina and Jiang, Li and Kosaraju, Vineet and Krueger, Gretchen and Navarro, Gabriel and Power, Alethea and others},
  journal={arXiv preprint arXiv:2112.09332},
  year={2021}
}

@inproceedings{yao2022react,
  title={ReAct: Synergizing reasoning and acting in language models},
  author={Yao, Shunyu and Zhao, Jeffrey and Yu, Dian and Peng, Shixiang and Narasimhan, Karthik and Cambria, Erik and Yang, Yiming},
  booktitle={Proceedings of the International Conference on Learning Representations (ICLR)},
  year={2023}
}

@article{schick2023toolformer,
  title={Toolformer: Language models can teach themselves to use tools},
  author={Schick, Timo and Dwivedi-Yu, Jane and Dess{\`i}, Roberto and Raileanu, Roberta and Lomeli, Maria and Scialom, Thomas and Sridhar, Abhinav and Beltagy, Iz and Launay, Julien and Schmidhuber, J{\"u}rgen and others},
  journal={arXiv preprint arXiv:2302.04761},
  year={2023}
}

@article{shinn2023reflexion,
  title={Reflexion: Language agents with verbal reinforcement learning},
  author={Shinn, Noah and Labash, Brendan and Gopinath, Dinesh},
  journal={arXiv preprint arXiv:2303.11366},
  year={2023}
}

@inproceedings{wei2022chain,
  title={Chain-of-thought prompting elicits reasoning in large language models},
  author={Wei, Jason and Wang, Xuezhi and Schuurmans, Dale and Bosma, Maarten and Ichter, Brian and Xia, Fei and Chi, Ed and Le, Quoc and Zhou, Denny},
  booktitle={Advances in Neural Information Processing Systems},
  year={2022}
}

@misc{openai2023gpt4,
  title={GPT-4 Technical Report},
  author={OpenAI},
  year={2023},
  howpublished={arXiv preprint arXiv:2303.08774}
}

@misc{shao_deepseekmath_2024,
	title = {{DeepSeekMath}: {Pushing} the {Limits} of {Mathematical} {Reasoning} in {Open} {Language} {Models}},
	shorttitle = {{DeepSeekMath}},
	url = {http://arxiv.org/abs/2402.03300},
	doi = {10.48550/arXiv.2402.03300},
	urldate = {2025-09-01},
	publisher = {arXiv},
	author = {Shao, Zhihong and Wang, Peiyi and Zhu, Qihao and Xu, Runxin and Song, Junxiao and Bi, Xiao and Zhang, Haowei and Zhang, Mingchuan and Li, Y. K. and Wu, Y. and Guo, Daya},
	month = apr,
	year = {2024},
	note = {arXiv:2402.03300 [cs]},
}

@misc{rafailov_direct_2024,
	title = {Direct {Preference} {Optimization}: {Your} {Language} {Model} is {Secretly} a {Reward} {Model}},
	shorttitle = {Direct {Preference} {Optimization}},
	url = {http://arxiv.org/abs/2305.18290},
	doi = {10.48550/arXiv.2305.18290},
	urldate = {2025-09-04},
	publisher = {arXiv},
	author = {Rafailov, Rafael and Sharma, Archit and Mitchell, Eric and Ermon, Stefano and Manning, Christopher D. and Finn, Chelsea},
	month = jul,
	year = {2024},
	note = {arXiv:2305.18290 [cs]},
}

@book{gamma1995design,
  author={Erich Gamma and Richard Helm and Ralph Johnson and John Vlissides},
  title={Design patterns: elements of reusable object-oriented software},
  publisher={Addison-Wesley Longman Publishing Co., Inc.},
  address={Boston, MA, United States},
  year={1995},
  isbn={978-0-201-63361-0},
  pages={395}
}

@misc{impala2018,
      title={IMPALA: Scalable Distributed Deep-RL with Importance Weighted Actor-Learner Architectures}, 
      author={Lasse Espeholt and Hubert Soyer and Remi Munos and Karen Simonyan and Volodymir Mnih and Tom Ward and Yotam Doron and Vlad Firoiu and Tim Harley and Iain Dunning and Shane Legg and Koray Kavukcuoglu},
      year={2018},
      eprint={1802.01561},
      archivePrefix={arXiv},
      primaryClass={cs.LG},
      url={https://arxiv.org/abs/1802.01561}, 
}

@misc{Huang2025Biomni,
  author       = {Huang, Kexin and Zhang, Serena and Wang, Hanchen and Qu, Yuanhao and Lu, Yingzhou and Roohani, Yusuf and Li, Ryan and Qiu, Lin and Li, Gavin and Zhang, Junze and Yin, Di and Marwaha, Shruti and Carter, Jennefer N. and Zhou, Xin and Wheeler, Matthew and Bernstein, Jonathan A. and Wang, Mengdi and He, Peng and Zhou, Jingtian and Snyder, Michael and Cong, Le and Regev, Aviv and Leskovec, Jure},
  title        = {Biomni: A General-Purpose Biomedical AI Agent},
  year         = {2025},
  howpublished = {bioRxiv preprint},
  doi          = {10.1101/2025.05.30.656746},
  url          = {https://www.biorxiv.org/content/early/2025/06/02/2025.05.30.656746}
}

@misc{luo2025agentlightningtrainai,
      title={Agent Lightning: Train ANY AI Agents with Reinforcement Learning}, 
      author={Xufang Luo and Yuge Zhang and Zhiyuan He and Zilong Wang and Siyun Zhao and Dongsheng Li and Luna K. Qiu and Yuqing Yang},
      year={2025},
      eprint={2508.03680},
      archivePrefix={arXiv},
      primaryClass={cs.AI},
      url={https://arxiv.org/abs/2508.03680}, 
}

@misc{hilton2025art,
  author = {Brad Hilton and Kyle Corbitt and David Corbitt and Saumya Gandhi and Angky William and Bohdan Kovalenskyi and Andie Jones},
  title = {ART: Agent Reinforcement Trainer},
  year = {2025},
  publisher = {GitHub},
  journal = {GitHub repository},
  howpublished = {\url{https://github.com/openpipe/art}}
}

@misc{langgraph2025,
  author       = {LangChain, Inc.},
  title        = {LangGraph},
  year         = {2025},
  howpublished = {\url{https://www.langchain.com/langgraph}},
}

@misc{gu2025surveyllmasajudge,
      title={A Survey on LLM-as-a-Judge}, 
      author={Jiawei Gu and Xuhui Jiang and Zhichao Shi and Hexiang Tan and Xuehao Zhai and Chengjin Xu and Wei Li and Yinghan Shen and Shengjie Ma and Honghao Liu and Saizhuo Wang and Kun Zhang and Yuanzhuo Wang and Wen Gao and Lionel Ni and Jian Guo},
      year={2025},
      eprint={2411.15594},
      archivePrefix={arXiv},
      primaryClass={cs.CL},
      url={https://arxiv.org/abs/2411.15594}, 
}

@misc{zheng2023judging,
      title={Judging LLM-as-a-Judge with MT-Bench and Chatbot Arena}, 
      author={Lianmin Zheng and Wei-Lin Chiang and Ying Sheng and Siyuan Zhuang and Zhanghao Wu and Yonghao Zhuang and Zi Lin and Zhuohan Li and Dacheng Li and Eric P. Xing and Hao Zhang and Joseph E. Gonzalez and Ion Stoica},
      year={2023},
      eprint={2306.05685},
      archivePrefix={arXiv},
      primaryClass={cs.CL},
      url={https://arxiv.org/abs/2306.05685}, 
}

@misc{liu2023geval,
      title={G-Eval: NLG Evaluation using GPT-4 with Better Human Alignment}, 
      author={Yang Liu and Dan Iter and Yichong Xu and Shuohang Wang and Ruochen Xu and Chenguang Zhu},
      year={2023},
      eprint={2303.16634},
      archivePrefix={arXiv},
      primaryClass={cs.CL},
      url={https://arxiv.org/abs/2303.16634}, 
}

@misc{schulman2017proximal,
      title={Proximal Policy Optimization Algorithms}, 
      author={John Schulman and Filip Wolski and Prafulla Dhariwal and Alec Radford and Oleg Klimov},
      year={2017},
      eprint={1707.06347},
      archivePrefix={arXiv},
      primaryClass={cs.LG},
      url={https://arxiv.org/abs/1707.06347}, 
}

@misc{christiano2017deep,
      title={Deep reinforcement learning from human preferences}, 
      author={Paul Christiano and Jan Leike and Tom B. Brown and Miljan Martic and Shane Legg and Dario Amodei},
      year={2023},
      eprint={1706.03741},
      archivePrefix={arXiv},
      primaryClass={stat.ML},
      url={https://arxiv.org/abs/1706.03741}, 
}

@misc{wang2023self,
      title={Self-Instruct: Aligning Language Models with Self-Generated Instructions}, 
      author={Yizhong Wang and Yeganeh Kordi and Swaroop Mishra and Alisa Liu and Noah A. Smith and Daniel Khashabi and Hannaneh Hajishirzi},
      year={2023},
      eprint={2212.10560},
      archivePrefix={arXiv},
      primaryClass={cs.CL},
      url={https://arxiv.org/abs/2212.10560}, 
}

@misc{huang2022large,
      title={Large Language Models Can Self-Improve}, 
      author={Jiaxin Huang and Shixiang Shane Gu and Le Hou and Yuexin Wu and Xuezhi Wang and Hongkun Yu and Jiawei Han},
      year={2022},
      eprint={2210.11610},
      archivePrefix={arXiv},
      primaryClass={cs.CL},
      url={https://arxiv.org/abs/2210.11610}, 
}

@misc{hinton2015distilling,
      title={Distilling the Knowledge in a Neural Network}, 
      author={Geoffrey Hinton and Oriol Vinyals and Jeff Dean},
      year={2015},
      eprint={1503.02531},
      archivePrefix={arXiv},
      primaryClass={stat.ML},
      url={https://arxiv.org/abs/1503.02531}, 
}

@misc{sanh2020distilbert,
      title={DistilBERT, a distilled version of BERT: smaller, faster, cheaper and lighter}, 
      author={Victor Sanh and Lysandre Debut and Julien Chaumond and Thomas Wolf},
      year={2020},
      eprint={1910.01108},
      archivePrefix={arXiv},
      primaryClass={cs.CL},
      url={https://arxiv.org/abs/1910.01108}, 
}

@misc{qin2023toolllm,
      title={ToolLLM: Facilitating Large Language Models to Master 16000+ Real-world APIs}, 
      author={Yujia Qin and Shihao Liang and Yining Ye and Kunlun Zhu and Lan Yan and Yaxi Lu and Yankai Lin and Xin Cong and Xiangru Tang and Bill Qian and Sihan Zhao and Lauren Hong and Runchu Tian and Ruobing Xie and Jie Zhou and Mark Gerstein and Dahai Li and Zhiyuan Liu and Maosong Sun},
      year={2023},
      eprint={2307.16789},
      archivePrefix={arXiv},
      primaryClass={cs.AI},
      url={https://arxiv.org/abs/2307.16789}, 
}

@misc{micikevicius2018mixed,
      title={Mixed Precision Training}, 
      author={Paulius Micikevicius and Sharan Narang and Jonah Alben and Gregory Diamos and Erich Elsen and David Garcia and Boris Ginsburg and Michael Houston and Oleksii Kuchaiev and Ganesh Venkatesh and Hao Wu},
      year={2018},
      eprint={1710.03740},
      archivePrefix={arXiv},
      primaryClass={cs.AI},
      url={https://arxiv.org/abs/1710.03740}, 
}

@misc{hu2021lora,
      title={LoRA: Low-Rank Adaptation of Large Language Models}, 
      author={Edward J. Hu and Yelong Shen and Phillip Wallis and Zeyuan Allen-Zhu and Yuanzhi Li and Shean Wang and Lu Wang and Weizhu Chen},
      year={2021},
      eprint={2106.09685},
      archivePrefix={arXiv},
      primaryClass={cs.CL},
      url={https://arxiv.org/abs/2106.09685}, 
}

@misc{dettmers2023qlora,
      title={QLoRA: Efficient Finetuning of Quantized LLMs}, 
      author={Tim Dettmers and Artidoro Pagnoni and Ari Holtzman and Luke Zettlemoyer},
      year={2023},
      eprint={2305.14314},
      archivePrefix={arXiv},
      primaryClass={cs.LG},
      url={https://arxiv.org/abs/2305.14314}, 
}

@misc{unsloth,
  author       = {Han, Daniel and Han, Michael},
  title        = {Unsloth},
  year         = {2023},
  howpublished = {\url{https://github.com/unslothai/unsloth}},
  note         = {GitHub repository}
}

@InProceedings{klimt2004enron,
    author="Klimt, Bryan
    and Yang, Yiming",
    editor="Boulicaut, Jean-Fran{\c{c}}ois
    and Esposito, Floriana
    and Giannotti, Fosca
    and Pedreschi, Dino",
    title="The Enron Corpus: A New Dataset for Email Classification Research",
    booktitle="Machine Learning: ECML 2004",
    year="2004",
    publisher="Springer Berlin Heidelberg",
    address="Berlin, Heidelberg",
    pages="217--226",
    series="Lecture Notes in Computer Science",
    volume="3201",
    isbn="978-3-540-30115-8",
    doi="10.1007/978-3-540-30115-8_22"
}

@misc{corbt_enron_questions_2025,
  author       = {Corbitt, Kyle},
  title        = {Enron Emails Sample Questions},
  year         = {2025},
  howpublished = {Hugging Face Datasets},
  url          = {https://huggingface.co/datasets/corbt/enron_emails_sample_questions}
}

@misc{arte_blog_post,
  author       = {Corbitt, Kyle},
  title        = {ART·E: How We Built an Email Research Agent That Beats o3},
  year         = {2025},
  month        = {April},
  howpublished = {OpenPipe Blog},
  url          = {https://openpipe.ai/blog/art-e-mail-agent}
}

\clearpage
\appendix

\section*{Appendix}
\addcontentsline{toc}{section}{Appendix}

\section{Implementation Details}
\label{sec:impl-details}

\subsection{Core API for the Email Search Agent}
\label{sec:core-api}

Listing~\ref{lst:core_api_email} demonstrates the core API used to set up the Email Search Agent experiment described in Section~\ref{sec:experiments}. It shows the agent definition and the \textbf{Brain} initialization. Crucially, it utilizes a custom reward function, \texttt{reward\_art\_style\_judge}, which was specifically designed to replicate the direct-assessment logic of the ART$\cdot$E project for a fair comparison of agent performance. This highlights ToolBrain's flexibility in accommodating highly specialized, user-defined reward signals.

\begin{lstlisting}[language=Python, caption={Core API setup for the Email Search Agent experiment.}, label={lst:core_api_email}]
from smolagents import tool, CodeAgent
from toolbrain.models import UnslothModel
from toolbrain import Brain
from my_project import custom_rewards # Custom reward functions

# 1. Define the agent's tools
@tool
def search_emails(keywords: list[str]) -> list[dict]:
    """Searches the email DB for given keywords."""
    # ... logic to query the email database ...
    return email_db.search(keywords)

# 2. Define the agent using the UnslothModel wrapper
email_search_agent = CodeAgent(
    model=UnslothModel(model_id="Qwen/Qwen2.5-7B-Instruct"),
    tools=[search_emails, read_email]
)

# 3. Initialize the Brain with the custom, ART-style judge
brain = Brain(
    agent=email_search_agent,
    algorithm="GRPO",
    reward_func=custom_rewards.reward_art_style_judge,
    reward_kwargs={"judge_model": "gpt-4o-mini"},
    learning_rate=3e-5
)

# 4. Start the training process
brain.train(dataset=email_questions_dataset)
\end{lstlisting}

\subsection{The Execution Trace: Core Data Structures}

The \texttt{Execution Trace} is the standardized, high-fidelity "source of truth" for all learning in ToolBrain. It is a list of \texttt{Turn} objects, where each \texttt{Turn} captures a complete interaction cycle. This data structure is generated by the internal \textbf{Agent Adapter}, which acts as an interpreter, translating an agent's framework-specific memory into this universal format. Listing~\ref{lst:trace_def} shows a simplified definition of these core structures, which are crucial for enabling accurate reward computation and RL training.

\begin{lstlisting}[language=Python, caption={A simplified definition of the core data structures.}, label={lst:trace_def}]
class Turn(TypedDict):
    prompt_for_model: str
    model_completion: str
    parsed_completion: ParsedCompletion
    tool_output: Optional[str]

class ParsedCompletion(TypedDict):
    thought: Optional[str]
    tool_code: Optional[str]
    final_answer: Optional[str]
\end{lstlisting}

\section{Reward Design}
\label{sec:reward-appendix}

ToolBrain supports two primary paradigms for reward design: simple, user-defined Python functions for straightforward tasks, and powerful LLM-based judges for more complex, nuanced evaluations.

\subsection{User-Defined Heuristic Rewards}

For tasks with clear, objective success criteria, users can provide any Python callable as a reward function. This function receives the full execution trace and returns a scalar score. Listing~\ref{lst:reward_api} shows an example that rewards efficiency.

\begin{lstlisting}[language=Python, caption={A simple user-defined reward function.}, label={lst:reward_api}]
from toolbrain.core_types import Trace

def reward_step_efficiency(trace: Trace, **kwargs) -> float:
    """Rewards higher for shorter traces."""
    max_turns = int(kwargs.get("max_turns", 5))
    num_turns = len(trace)
    
    if num_turns <= max_turns:
        return 1.0
    penalty = (num_turns - max_turns) * 0.1
    return max(0.0, 1.0 - penalty)
\end{lstlisting}

\subsection{LLM-as-a-Judge Reward Function}

For complex tasks without a clear ground-truth, ToolBrain provides a built-in, ranking-based LLM-as-a-judge. This function operates on a batch of traces and does not require a \texttt{gold\_answer}, making it ideal for unsupervised learning scenarios. As shown in Listing~\ref{lst:llm_judge_api}, the user simply passes this function to the \textbf{Brain}. The framework then automatically handles the process of collecting multiple traces, prompting a judge model to rank them, and converting these ranks into scalar rewards. While our main experiment (Section~\ref{sec:experiments}) used a custom direct-assessment judge for methodological consistency with prior work, this built-in approach is the recommended, general-purpose solution.

\begin{lstlisting}[language=Python, caption={Using ToolBrain's built-in, ranking-based LLM judge.}, label={lst:llm_judge_api}]
from toolbrain.rewards import reward_llm_judge_via_ranking

brain = Brain(
    agent=my_agent,
    algorithm="GRPO",
    reward_func=reward_llm_judge_via_ranking,
    reward_kwargs={"judge_model": "gemini/gemini-1.5-flash"}
)
\end{lstlisting}

\section{Learning Algorithms}
\label{sec:rl-algorithms}

We provide here the GRPO and DPO training procedures referenced in
Section~\ref{sec:key-features}.

\begin{algorithm*}[t]
\caption{GRPO training for a single query $q$}
\label{alg:grpo}
\begin{algorithmic}[1]
\REQUIRE Policy model $\pi_\theta$, reward function $R$, group size $G$,
hyperparameters $\epsilon, \beta$
\STATE For $i = 1,\dots,G$, run the agent to obtain a Trace $\tau_i$
\STATE Compute a scalar reward $r_i = R(\tau_i)$ for each trace
\STATE Compute group-normalized advantage
\[
\hat{A}_i = \frac{r_i - \operatorname{mean}(\{r_j\}_{j=1}^G)}
                 {\operatorname{std}(\{r_j\}_{j=1}^G)}
\]
\STATE Assemble the GRPO loss
\[
\mathcal{L}_{\text{GRPO}}(\theta) = - \frac{1}{G}\sum_{i=1}^{G}
\frac{1}{|o_i|}\sum_{t=1}^{|o_i|}
\Big[
\min(\rho_{i,t}\hat A_{i,t},
\operatorname{clip}(\rho_{i,t},1-\epsilon,1+\epsilon)\hat A_{i,t})
- \beta D_{\mathrm{KL}}(\pi_\theta \Vert \pi_{\mathrm{ref}})
\Big]
\]
\STATE Update $\theta \leftarrow \theta - \eta \,\nabla_\theta \mathcal{L}_{\text{GRPO}}(\theta)$
\end{algorithmic}
\end{algorithm*}

\begin{algorithm*}[t]
\caption{DPO training for a single query $q$}
\label{alg:dpo}
\begin{algorithmic}[1]
\REQUIRE Policy $\pi_\theta$, reference policy $\pi_{\text{ref}}$,
group size $G$, hyperparameter $\beta$
\STATE For $i = 1,\dots,G$, run the agent to obtain a Trace $\tau_i$
\STATE Compute rewards and sample a preferred $y_w$ and dispreferred
$y_\ell$ for each query
\STATE Define
\[
r_\theta(y \mid x) =
\log \frac{\pi_\theta(y \mid x)}{\pi_{\text{ref}}(y \mid x)}
\]
\STATE Assemble the DPO loss
\[
\mathcal{L}_{\text{DPO}}(\theta)
= - \log \sigma\left(
\beta \big( r_\theta(y_w \mid x) - r_\theta(y_\ell \mid x) \big)
\right)
\]
\STATE Update $\theta \leftarrow \theta - \eta \,\nabla_\theta \mathcal{L}_{\text{DPO}}(\theta)$
\end{algorithmic}
\end{algorithm*}

\section{Zero-Learn Task Generation Details}
\label{sec:zero-learn}

This section expands on the Zero-Learn mechanism briefly described in
Section~\ref{sec:key-features}.

\subsection{Generation API}

\begin{lstlisting}[caption={Generating training examples with Brain and a Qwen model.}, label={lst:generate_examples_api},language={Python}]
from smolagents import CodeAgent
from toolbrain import Brain, get_transformer_model

agent = CodeAgent(
    model=TransformersModel(model_id="Qwen/Qwen2.5-0.5B-Instruct"),
    tools=[
        calculate_compound_interest,
        calculate_loan_payment,
        calculate_cagr,
        calculate_npv
    ],
)

brain = Brain(
    agent=agent,
    algorithm="GRPO"
    # reward_func is not needed for generation
)

generated_examples = brain.generate_training_examples(
    task_description="Generate tasks to learn to use simple finance tools.",
    num_examples=100,
    min_tool_calls=2,
    max_words=80,
    self_rank=True
)
\end{lstlisting}

\subsection{Example Query Categories}

We observed three main categories of generated queries:

\textbf{(i) Executable tool calls}
\begin{lstlisting}[language={},numbers=none]
"Calculate Loan Payment with annual rate of 5%, 7 years, principal of $10,000."
"Calculate Compound Interest: Principal = 1000, Rate = 0.05,
 Times Compounded = 12, Years = 10"
"What is the compound interest on $10,000 at an annual interest
 rate of 5% for 3 years?"
\end{lstlisting}

\textbf{(ii) Formula or explanatory requests}
\begin{lstlisting}[language={},numbers=none]
"What is the formula for calculating compound interest?"
"What is the formula to calculate the future value of an investment?"
\end{lstlisting}

\textbf{(iii) Out-of-scope or noisy queries}
\begin{lstlisting}[language={},numbers=none]
"Calculate the total cost of a car purchase including insurance
 and maintenance over 5 years ..."
"Calculate Compound Interest on $10,000 for 3 years at an annual
 rate of 5%, then convert this amount to USD using the current
 exchange rate and compute the NPV."
\end{lstlisting}

Out of 100 generated queries, approximately 63\% were directly
executable, 27\% were formula or explanatory, and 10\% were noisy or out-of-scope. These statistics informed our filtering and rewriting strategy.

\section{Knowledge Distillation Pipeline}
\label{sec:distillation}

\subsection{Algorithm}

\begin{algorithm*}[t]
\caption{ToolBrain Distillation Pipeline}
\label{alg:distill}
\begin{algorithmic}[1]
\STATE \textbf{Input:} Teacher model $\pi_T$, student brain $B_S$,
tool function $\mathcal{T}$, query set $q$
\STATE \textbf{Parameters:} $N = 100$ traces, quality threshold $\rho$
\IF{cached traces exist}
    \STATE Load $(\{\tau_i\}, \{x_i\}, \{r_i\})$ from disk
\ELSE
    \STATE Initialize teacher agent with $\pi_T$ and tool $\mathcal{T}$
    \FOR{$i = 1$ to $N$}
        \STATE Execute teacher agent on query $q$
        \STATE Collect trace $\tau_i$, RL input $x_i$, reward $r_i$
    \ENDFOR
    \STATE Cache $(\{\tau_i\}, \{x_i\}, \{r_i\})$ to disk
\ENDIF
\STATE Filter high-quality traces: $\mathcal{F} = \{x_i \mid r_i > \rho\}$
\IF{$|\mathcal{F}| > 0$}
    \STATE Train student $\pi_S$ with cross-entropy on $\mathcal{F}$:
    \[
    \mathcal{L}_{\text{distill}}(\theta) = -\frac{1}{|\mathcal{F}|}
      \sum_{x \in \mathcal{F}} \sum_{t=1}^{|y|}
      \log \pi_S(y_t | x, y_{<t})
    \]
\ENDIF
\STATE \textbf{Return:} Pre-trained student model
\end{algorithmic}
\end{algorithm*}

\subsection{Example Usage}
The distillation pipeline is highly flexible and can be used in two primary ways depending on the task's complexity.

\subsubsection{Distillation as a Standalone Method}
For simpler, single-step tasks where the goal is to learn a direct mapping from query to a tool call, distillation can be used as a complete, standalone training method. This efficient approach is what we employed for our supplementary experiments in Section~\ref{sec:experiments}. Listing~\ref{lst:distillation-standalone} demonstrates this usage.

\begin{lstlisting}[language=Python, caption=Distillation as a Standalone Training Method, label={lst:distillation-standalone}]
from toolbrain import Brain
from smolagents import CodeAgent, TransformersModel

# Define the student agent
student_agent = CodeAgent(tools=[my_tool], ...)

# Initialize the Brain
brain = Brain(agent=student_agent, reward_func=my_reward_func)

# Run the complete distillation pipeline
brain.distill(
    dataset=training_tasks,
    teacher_model_id="Qwen/Qwen2.5-7B-Instruct",
    num_traces=len(training_tasks)
)

\end{lstlisting}

\subsubsection{Distillation as a Warm-Up for RL}
For more complex, multi-step tasks that benefit from further exploration and policy refinement, distillation serves as a powerful warm-up phase before reinforcement learning. This two-stage process, demonstrated in our codebase's \texttt{08\_distillation.py} example, first initializes the agent with high-quality behaviors and then uses RL to improve upon them. Listing~\ref{lst:distillation-warmup} illustrates this workflow.

\begin{lstlisting}[language=Python, caption=Distillation as a Warm-Up for RL, label={lst:distillation-warmup}]
from toolbrain import Brain
from smolagents import CodeAgent, TransformersModel

student_model = TransformersModel("Qwen/Qwen2.5-0.5B-Instruct")
student_agent = CodeAgent(tools=[my_tool], model=student_model)

brain = Brain(
    agent=student_agent,
    reward_func=my_reward_function,
    algorithm="GRPO"
)

brain.distill(
    dataset=training_tasks,
    teacher_model_id="Qwen/Qwen2.5-7B-Instruct"
)

brain.train(training_tasks, num_iterations=5)
\end{lstlisting}

\section{Tool Management and Efficiency}
\label{sec:tools-efficiency}

\subsection{Tool Retriever}

\begin{lstlisting}[language=Python, caption=Tool Retriever Usage,
label={lst:tool_retriever}]
from smolagents import CodeAgent
from toolbrain import Brain
from toolbrain.retriever import ToolRetriever
from openai import OpenAI

# Assume math_tools are defined with @tool decorator
all_math_tools = [add, multiply, divide, subtract]

agent = CodeAgent(
    model="Qwen/Qwen2.5-0.5B-Instruct",
    tools=all_math_tools
)

client_instance = OpenAI(api_key=...).chat.completions.create

retriever = ToolRetriever(
    llm_model="gpt-4o-mini",
    llm_instance=client_instance,
    retrieval_topic="mathematics",
    retrieval_guidelines="Select only necessary tools..."
)

brain = Brain(
    agent=agent,
    algorithm="GRPO",
    tool_retriever=retriever
)
\end{lstlisting}

\subsection{Training Optimizations}
\label{sec:training-opts-appendix}

ToolBrain exposes complex training optimizations through simple, high-level parameters. Listing~\ref{lst:training_opts} demonstrates two common methods for reducing memory usage: Option 1 shows standard mixed-precision training (\texttt{fp16}), while Option 2 demonstrates how to enable QLoRA by using the \texttt{BitsAndBytesConfig} from the \texttt{bitsandbytes} library to configure 4-bit quantization. This API design pattern extends to other optimizations as well; for instance, a PEFT \texttt{LoraConfig} can be passed similarly via \texttt{model\_kwargs}, and specialized models like \texttt{UnslothModel} can be used as a drop-in replacement for \texttt{TransformersModel} to leverage further acceleration.

\begin{lstlisting}[language=Python, caption=Enabling Training Optimizations,
label={lst:training_opts}]
from smolagents import CodeAgent, TransformersModel
from toolbrain import Brain
from transformers import BitsAndBytesConfig
import torch

# --- Option 1: Mixed-Precision (FP16) ---

# Load the model with float16 data type
model_fp16 = TransformersModel(
    model_id="Qwen/Qwen2.5-0.5B-Instruct",
    torch_dtype=torch.float16
)
agent_fp16 = CodeAgent(model=model_fp16, tools=...)

# Enable the fp16 optimizer in Brain
brain_fp16 = Brain(
    agent_fp16,
    algorithm="GRPO",
    fp16=True
)

# --- Option 2: 4-bit Quantization (QLoRA) ---

# Define the 4-bit quantization configuration
nf4_config = BitsAndBytesConfig(
    load_in_4bit=True,
    bnb_4bit_quant_type="nf4"
)

# Pass the config to the model via model_kwargs
model_qlora = TransformersModel(
    model_id="Qwen/Qwen2.5-0.5B-Instruct",
    model_kwargs={"quantization_config": nf4_config}
)
agent_qlora = CodeAgent(model=model_qlora, tools=...)

# Enable the corresponding 8-bit optimizer in Brain
brain_qlora = Brain(
    agent_qlora,
    algorithm="GRPO",
    use_bitsandbytes=True
)
\end{lstlisting}

\end{document}